\documentclass{article}

\usepackage{iclr2026_conference}
\iclrfinalcopy
\usepackage{lmodern}
\usepackage{float}
\renewcommand{\cite}{\citep} %

\usepackage[utf8]{inputenc} %
\usepackage[T1]{fontenc}    %
\usepackage{hyperref}       %
\usepackage{url}            %
\usepackage{booktabs}       %
\usepackage{amsfonts}       %
\usepackage{nicefrac}       %
\usepackage{microtype}      %
\usepackage{xcolor}         %

\usepackage[pdftex]{graphicx}
\usepackage{siunitx}

\title{Fine-tuning language encoding models on slow fMRI improves prediction for fast ECoG}

\author{%
  Aditya R. Vaidya \\
  Department of Computer Science\\
  The University of Texas at Austin, USA \\
  \texttt{avaidya@utexas.edu} \\
  \And
  Richard J. Antonello \\
  Zuckerman Mind Brain Behavior Institute \\
  Columbia University, USA \\
  \texttt{rja2163@columbia.edu} \\
  \AND
  Alexander G. Huth \\
  Departments of Neuroscience and Statistics \\
  University of California, Berkeley, USA \\
  \texttt{alex.huth@berkeley.edu} \\
}

\begin{document}

\maketitle

\begin{abstract}
Neuroscientists have recently turned to intracranial brain recording methods, like electrocorticography (ECoG), for human experiments because of the fine spatial and temporal resolution that they afford.
Models trained on this data, however, are fundamentally restricted by the patient populations that can receive the implants necessary for recording.
We propose using non-invasive fMRI to bridge the gap in training data.
Using spoken language representations fine-tuned on fMRI, we build encoding models of ECoG.
These representations showed improved prediction performance in ECoG, even though the temporal resolution of fMRI is two orders of magnitude worse.
Prediction improved in frequency bands well beyond what is directly measured in fMRI.
Next, to test the procedure's generalization ability, we fine-tuned models on fMRI responses that were temporally downsampled by a factor of 2.
Despite the loss in resolution, these models were able to predict fMRI and ECoG responses at levels comparable to the original fMRI-tuned models.
Finally, we showed that ECoG performance steadily scales with the amount of fMRI-tuning data.
Our results show that ``slow'' data like fMRI can be a valuable resource for building better models of ``fast'' brain data like ECoG.
In the future, integrating across multiple recording methods may further improve performance in other applications, like decoding.
\end{abstract}

\section{Introduction}

Many neuroscience experiments use data recorded from intracranial electrodes that have been surgically implanted in human patients as part of clinical treatment. In electrocorticography (ECoG) the electrodes are arranged in grids that are applied directly to the surface of the brain. Electrodes in such close proximity to the brain tissue record neural activity with high temporal and spatial precision, enabling the construction of detailed encoding models that predict brain activity elicited by a stimulus. These encoding models can then be used to answer neuroscience research questions~\cite{mesgaraniPhoneticFeatureEncoding2014,keshishianParallelHierarchicalEncoding2026} or in downstream applications like brain-computer interfaces (BCIs)~\cite{tangSemanticReconstructionContinuous2023,littlejohnStreamingBraintovoiceNeuroprosthesis2025}.

The downside of intracranial recordings is that datasets are small and rare.
ECoG electrodes are only implanted when clinically necessary, and are typically removed within a few days~\cite{changLargeScaleHumanBasedMesoscopic2015,mercierAdvancesHumanIntracranial2022}.
The number of electrodes and their placement also differs from patient to patient, depending on their clinical needs.
This severely limits the amount of data collected from each individual, making it challenging to build elaborate encoding models.
In contrast, non-invasive measurements like functional magnetic resonance imaging (fMRI)---while having much lower temporal resolution than ECoG---are easy to acquire repeatedly and have whole-brain coverage.
Can plentiful fMRI data be exploited to improve encoding models for intracranial data?

Modern encoding models are typically based on pretrained deep neural networks. Language encoding models often use neural network language models pretrained on large text corpora~\cite{jainIncorporatingContextLanguage2018,caucheteuxBrainsAlgorithmsPartially2022} or speech models pretrained on audio corpora~\cite{vaidyaSelfSupervisedModelsAudio2022,milletRealisticModelSpeech2022,tuckuteManyNotAll2023}. These networks have mostly been used as frozen feature extractors~\cite{kellTaskOptimizedNeuralNetwork2018,ootaSpeechLanguageModels2024}, but in more recent ``brain-tuning'' studies the networks are also fine-tuned on brain data to improve encoding performance~\cite{moussaImprovingSemanticUnderstanding2024a,negiBrainInformedFineTuningImproved2025}. fMRI-tuned models increase prediction performance relative to pretrained models~\cite{vattikondaBrainWavLMFinetuningSpeech2025}, and also generalize to fMRI data from new subjects or different brain areas~\cite{moussaBraintuningImprovesGeneralizability2025}.
While the limited size of ECoG datasets make fine-tuning challenging, it may be possible that fMRI-tuned models, by virtue of learning more brain-like representations, generalize to ECoG. This would enable ECoG encoding model performance to scale with available fMRI datasets.

The idea that ECoG encoding models could benefit from fine-tuning on fMRI data is surprising but not impossible. ECoG measures electrical activity in the brain that varies at the scale of milliseconds, while fMRI measures blood flow that varies over seconds---at least 2 orders of magnitude slower. Yet despite its low sampling rate, fMRI signals are sensitive to fast timescale stimulus properties. In auditory cortex, for example, fMRI responses can depend upon temporal modulations in sound that are much faster than the fMRI signal itself \cite{overathCorticalAnalysisSpeechspecific2015,schonwiesnerSpectrotemporalModulationTransfer2009}. More recent studies have shown that auditory cortex fMRI responses to speech are well-modeled using networks like HuBERT~\cite{hsuHuBERTSelfSupervisedSpeech2021}, WavLM~\cite{chenWavLMLargeScaleSelfSupervised2021}, or Whisper~\cite{radfordRobustSpeechRecognition2022}. These models' fMRI prediction performance are improved by fine-tuning on fMRI data~\cite{moussaImprovingSemanticUnderstanding2024a,vattikondaBrainWavLMFinetuningSpeech2025} And the same networks can also very effectively predict ECoG responses to speech~\cite{liDissectingNeuralComputations2022}. These results all support the possibility that fMRI-tuned models could improve ECoG prediction performance.

\begin{figure}[t]
	\centering
	\includegraphics[width=1\linewidth]{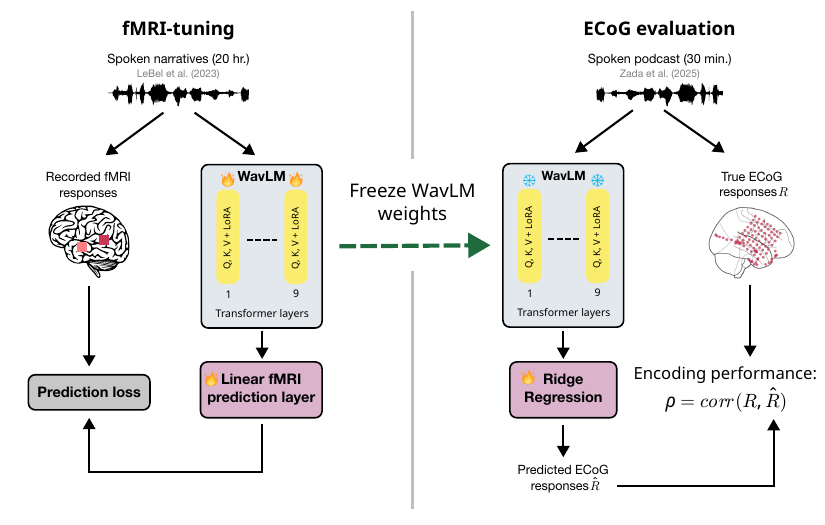}
	\caption{\textbf{fMRI-to-ECoG transfer via fMRI-tuning.} We fine-tune the 9th layer of a deep speech representation model, WavLM Base+~\cite{chenWavLMLargeScaleSelfSupervised2021}, to predict fMRI responses (measured at 0.5 Hz) to spoken language. We then freeze the weights of the WavLM model and use its representations to build linearized encoding models of ECoG responses (measured at 20 Hz) to speech from a separate dataset. Successfully performing this task requires learning representations that are useful across brain recording methods and robust to new subjects and stimuli.}
	\label{fig:intro}
\end{figure}

In this work, we demonstrate that encoding models fine-tuned on fMRI can generalize to new subjects, stimuli, and recording methods (Figure~\ref{fig:intro}).
Models fine-tuned on fMRI improved prediction performance in ECoG, despite the difference in temporal resolution between the two methods.
We stress test the ``temporal resolution generalization'' of this procedure by fine-tuning on downsampled fMRI responses. Despite a sampling rate of only 0.25 Hz, fine-tuning on these responses still yielded a significant improvement in ECoG prediction performance compared to the pretrained model. We show that this slow-to-fast generalization even applies within fMRI data; models fine-tuned on downsampled fMRI responses were able to predict the original fMRI responses better than the pretrained model.
Finally, we show that ECoG performance steadily increases with the amount of fMRI fine-tuning data, demonstrating that scaling fMRI language datasets could benefit ECoG models for applications in neuroscience or BCI, despite the vast difference in temporal resolution.

\section{Data and methods}
\label{sec:methods}

\subsection{fMRI data}

From the public dataset released by LeBel et al. (2023)~\cite{lebelNaturalLanguageFMRI2023} and Tang et al. (2023)~\cite{tangSemanticReconstructionContinuous2023}, we used pre-processed fMRI data from 3 participants who were scanned while listening to 94--103 naturally spoken narrative stories (17.8 h--19.7 h hours per participant). These stories capture a range of responses that may be easier to capture in whole-brain recordings (like fMRI) than in more limited intracranial datasets. Three of the stories were presented to the subjects multiple times: two (``fromboyhoodtofatherhood'' and ``onapproachtopluto'') were seen five times each, and one (``wheretheressmoke'') ten times. We averaged across the responses of the repeated presentations to reduce measurement noise.

\subsection{Speech encoding models}
\label{sec:speech-encmodels}

In this work, we build linearized speech encoding models that aim to estimate the response $R$ to a stimulus $S$ as:
\begin{equation}
    \hat{R}_t = f(S_t;\theta) \beta
\end{equation}
where $f: \mathbb{S} \rightarrow \mathbb{R}^F $ is a non-linear transform of the stimulus $S$ at time $t$ with parameters $\theta$, and $\beta \in \mathbb{R}^{F \times C}$ is a linear projection of the $F$ features to each of $C$ channels of the brain response. 
Response channels are voxels in fMRI and electrodes in ECoG.
When possible, encoding models use a finite impulse response (FIR) structure to capture temporal properties of the response. For example, fMRI responses derive from the blood-oxygen-level-dependent (BOLD) signal which, after an impulse of neural activity, rises to a peak over the course of 3-4 s and then falls back to baseline over another 4-6 s~\cite{naselarisEncodingDecodingFMRI2011}. To capture this behavior, our fMRI encoding models use concatenated stimulus features from several timepoints ($t-4$, $t-3$, $t-2$, and $t-1$) to predict the responses at time $t$. For an FIR model with $d$ delays, this expands the feature space to $d\cdot F$ dimensions.
In ECoG, responses are also delayed by upstream neural processing, often spanning 200-800 ms after a stimulus transient~\cite{hullettHumanSuperiorTemporal2016,hamiltonSpatialMapOnset2018}. Because the sampling rate of ECoG data is also much higher, capturing this temporal response could require $d=20$ or more. This quickly becomes computationally intractable for large feature spaces. To circumvent this issue, we followed earlier studies~\cite{goldsteinUnifiedAcoustictospeechtolanguageEmbedding2025,zadaPodcastECoGDataset2025} and fit ECoG encoding models that use only a single delay hyperparameter $\tau$, yielding $\hat{R}_t = f(S_{t-\tau}; \theta) \beta$. 

Here, we parametrize the non-linear stimulus transform $f$ using WavLM~\cite{chenWavLMLargeScaleSelfSupervised2021}, a neural network that operates on the waveform of the stimulus.
Layer 9 has been previously shown to have the highest encoding performance in fMRI~\cite{antonelloScalingLawsLanguage2023}, so we only extract features from this layer.
Following previous work, we extract features by sliding a 4 s window over the waveform, feeding the windowed stimulus into the model, and saving the hidden state of the final token of the layer.

For modeling fMRI, we extract features with a 0.25 s stride, and the result is downsampled to 0.5 Hz, the sampling rate of the fMRI responses. 
For modeling ECoG, we extract features with a 0.05 s stride, resulting in features at 20 Hz, the sampling rate we use for the ECoG responses.

\subsection{fMRI-tuning of audio models}
\label{sec:methods-fmri-tuning}

To induce a brain-like bias in the speech representations, we fine-tune the underlying WavLM model to predict fMRI responses in a procedure we call ``fMRI-tuning''.
We adopt the procedure of Vattikonda et al. (2025)~\cite{vattikondaBrainWavLMFinetuningSpeech2025} to fine-tune separate WavLM models to predict the fMRI responses of three subjects from the fMRI dataset, starting from the WavLM Base+ checkpoint.
To reduce the likelihood of overfitting, we use rank-4 low-rank adaptation (LoRA~\cite{huLoRALowRankAdaptation2021}) on the $W^Q, W^K, Q^V$ matrices of each Transformer layer, and we constrain the final linear projection from WavLM to fMRI voxels to be rank-100.
We use the Adam optimizer~\cite{kingmaAdamMethodStochastic2017} with a learning rate of $5\times 10^{-4}$ to optimize the LoRA matrices and linear projection to minimize a spatial correlation loss.

We use two stories as a validation set (``fromboyhoodtofatherhood'' and ``onapproachtopluto'') and, as we do for the pretrained model, evaluate encoding performance on one test story (``wheretheressmoke'').
We fine-tuned each model for 30 epochs with a batch size of 10 TRs using the same feature extraction parameters as described in Section~\ref{sec:speech-encmodels}.
To select the best epoch, we evaluated encoding performance on the validation set.
Using the ridge parameters from the pretrained model, we fit ridge regression encoding models using features from each of the 30 epochs, and we selected the epoch with the best validation encoding performance.
For this epoch, we then re-ran cross-validation to select the best ridge parameters, and we evaluated its encoding performance on the unseen test story.

Fine-tuning one model took 30 hours on a single 48GB NVIDIA RTX A6000.

\subsection{ECoG data and evaluation}

We use the ``Podcast'' dataset~\cite{zadaPodcastECoGDataset2025} to evaluate the generalization of our fMRI-tuned models to intracranial data.
In this dataset, nine patients listened to a 30-minute podcast while electrical brain activity was recorded with electrocorticography (ECoG).
The dataset contains 1,268 electrodes across all patients, who are distinct from the participants in the fMRI dataset.

As is conventional for ECoG, we used the high-gamma power (provided by \cite{zadaPodcastECoGDataset2025}) in each ECoG electrode as the response of interest~\cite{mukamelCouplingNeuronalFiring2005,manningBroadbandShiftsLocal2009}. Power in the high-gamma band was computed as the analytic amplitude of the Hilbert transform of ECoG signals that had been band-pass filtered in the 70-200 Hz band. High-gamma power is thought to represent pooled activity across synchronously active neurons in the region near the electrode.
We further downsampled the high-gamma signals to 20 Hz to reduce computational burdens.

As mentioned earlier, the high temporal resolution of ECoG and high dimensionality of the speech representations make it computationally difficult to fit FIR encoding weights.
Instead, we offset the stimulus different amounts relative to the response, and build separate linear models for each lag with ridge regression. 81 lags are evenly spaced between -2 and +2 seconds. ECoG encoding performance is calculated with 4-fold cross-validation on the 30-minute stimulus.
The encoding performance for a model is that of the best performing lag.

Along with encoding performance (Pearson correlation coefficient) on the overall signal, we also evaluate performance within individual frequency bands by computing the power spectrum density (PSD) of the residual of the model's predictions.
The residual PSD for a model is chosen to be that of the lag with the highest encoding performance.

Fitting all ECoG encoding models for one fMRI-tuned model took 10 hours on a single NVIDIA RTX A6000.

\subsection{Signal-to-noise ratio of fMRI responses}
\label{sec:methods-snr}

fMRI data comprise both signal and noise. To tease the two apart one must use data collected while the same stimulus is played several times. Averaging the response timecourse across repeats gives an estimate of the ``signal'' portion of the data. Subtracting that average from each individual recording then gives an estimate of the ``noise'' part. Computing the variance of the signal and noise components then enables one to directly estimate the total signal to noise ratio (SNR). We estimate the SNR within each frequency band in a similar fashion: compute the power spectral densities (PSD) of the signal and noise components, and then compute the ratio at each frequency\cite{hsuQuantifyingVariabilityNeural2004a}. We use the 10 individual presentations of the test story (``wheretheressmoke'') to compute band-wise SNR in our fMRI data.

\section{Results}
\label{sec:results}

\begin{figure}[t]
    \centering
    \includegraphics[width=1\linewidth]{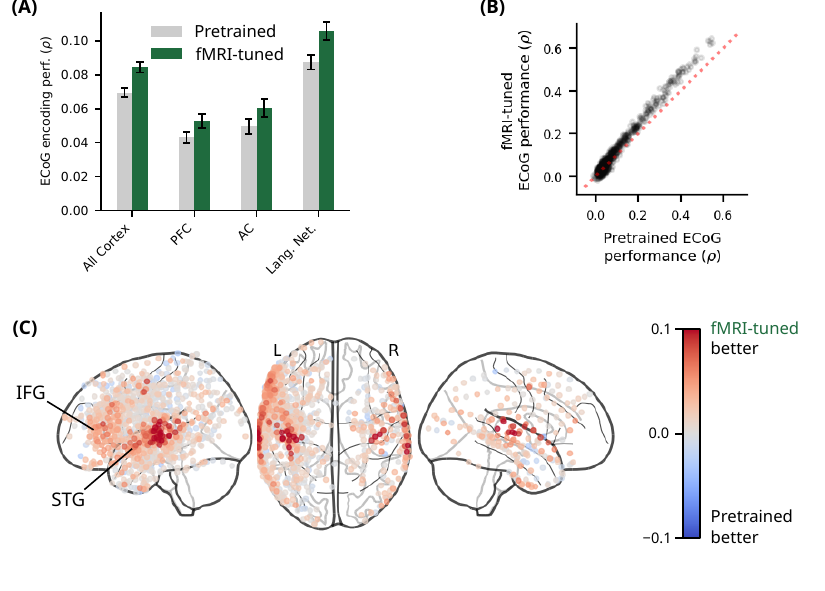}
    \caption{\textbf{fMRI-tuned models for predicting ECoG.} (\textbf{a}) ECoG encoding performance averaged across all electrodes within regions of interest (ROIs). Error bars show the standard error of the mean (SEM) across electrodes. The Destrieux atlas~\cite{destrieuxAutomaticParcellationHuman2010} was used to find electrodes in each ROI: pre-frontal cortex (PFC), primary auditory cortex (AC), and the language network as described in Lipkin et al. (2022)~\cite{lipkinProbabilisticAtlasLanguage2022}. fMRI-tuned models outperform pretrained models in every ROI ($p < \num{1e-8}$ for paired $t$-tests across electrodes within the ROI). (\textbf{b}) Encoding performance of each electrode using the pretrained vs. fMRI-tuned models. fMRI-tuned models consistently outperform pretrained models, especially for electrodes that were already well-modeled ($p < \num{1e-87}$ for a paired $t$-test across all electrodes). (\textbf{c}) Change in encoding performance for each electrode after fMRI-tuning, visualized on a brain atlas. Electrodes with the greatest improvement are found in bilateral auditory cortex. Smaller gains are seen in other components of the language network, including the superior temporal gyrus (STG) and inferior frontal gyrus (IFG). In Appendix~\ref{app:ecog-per-subj}, we separate the electrodes by subject (Figure~\ref{fig:app-ecog-per-subj}).}
    \label{fig:fmri2ecog_results}
\end{figure}

\subsection{Fine-tuning on fMRI improves ECoG performance}

Following Vattikonda et al. 2025~\cite{vattikondaBrainWavLMFinetuningSpeech2025}, we fine-tuned separate WavLM models on responses from each fMRI participant. Using features from these fine-tuned models, we fit new encoding models to ECoG responses from the ``Podcast'' dataset.

Averaged across cortex, encoding models fine-tuned on fMRI outperform the pretrained WavLM model when predicting ECoG (Figure~\ref{fig:fmri2ecog_results}a).
Improvements are most pronounced in electrodes within auditory cortex (AC) and the ``language network'', a set of areas hypothesized to be selective for language processing~\cite{lipkinProbabilisticAtlasLanguage2022}.
Electrodes that did not improve after fine-tuning had poor pretrained performance as well, suggesting their response is not stimulus-driven (Figure~\ref{fig:fmri2ecog_results}b).

In Figure~\ref{fig:fmri2ecog_results}c, we show the change in performance after fMRI-tuning for each electrode on an atlas brain map. Despite evaluating on a different recording modality, the cortical areas that improve in ECoG after fine-tuning largely overlap with previous fMRI-specific findings~\cite{moussaBraintuningImprovesGeneralizability2025,negiBrainInformedFineTuningImproved2025,vattikondaBrainWavLMFinetuningSpeech2025}.

\subsection{fMRI-tuning improves prediction in high-frequency bands of high-gamma}

Responses in auditory cortex (AC) tend to follow spectral features of the stimulus and can fluctuate quickly~\cite{hullettHumanSuperiorTemporal2016,norman-haignereMultiscaleTemporalIntegration2022}. %
Interestingly, electrodes in AC also showed the greatest improvements after fMRI-tuning.
This led us to ask what parts of the neural response were better predicted after fMRI-tuning.
We characterized the spectral properties of the model's performance by examining the power spectral density of the residuals from the encoding model predictions.
Because fMRI responses are temporally smooth, one might expect fMRI-tuning to only improve predictions for slowly-varying components of the ECoG response at or below the fMRI Nyquist rate (0.25 Hz).
However, these fMRI signals still ultimately derive from neural responses to rapidly-changing speech stimuli, which we know are well-modeled by the WavLM features.
Thus it is also possible that fine-tuning captures or accentuates brain-relevant information at much finer timescales, leading to improvements in ECoG prediction performance above 0.25 Hz.

\begin{figure}
    \centering
    \includegraphics[]{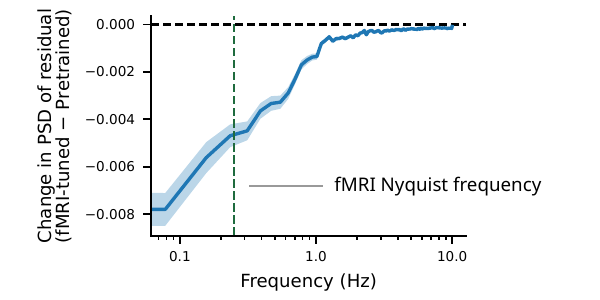}
    \caption{\textbf{Frequency-binned improvement of fMRI-tuned models.} Change in the power spectral density (PSD) of the ECoG encoding model residual after fMRI-tuning (lower is better). The shaded area shows the standard error across electrodes. The dotted green line at 0.25 Hz shows the Nyquist frequency of the fMRI responses. fMRI-tuning improves model fit (reduces the residual power) overall, with improvement both below and above the fMRI Nyquist frequency (two-sided $t$-test across electrodes between pretrained and fMRI-tuned PSD, $p<\num{5e-5}$ for every frequency). The power of the residual below 0.25 Hz decreases by $1.16\%$ ($p < \num{1e-23}$, two-sided $t$-test), and above 0.25 Hz the power decreases by $0.602\%$ ($p < \num{1e-31}$, two-sided $t$-test). This demonstrates that models fine-tuned on fMRI data can generalize and improve prediction performance of responses much faster than can be measured in fMRI.}
    \label{fig:spectrum}
\end{figure}

Surprisingly, we observed that fMRI-tuning improved the prediction of ECoG across the entire spectrum (Figure~\ref{fig:spectrum}). The biggest improvements were at frequencies below 0.25 Hz, but substantial and significant improvements were seen at frequencies up to 1 Hz. Even in the 1-10 Hz band we found small but consistent improvement from fMRI-tuning.
The power of the error reduced by $1.16\%$ below 0.25 Hz and above $0.602\%$ above it.
This result indicates that models tuned to predict fMRI data also predict ECoG at timescales that fMRI cannot resolve. 

\subsection{Models fine-tuned on downsampled fMRI generalize to faster responses}

In the previous sections we showed that fMRI-tuned models generalize to the higher temporal resolution of ECoG.
How far can we push this temporal generalization?
To further stress-test our approach we created a new dataset at even lower temporal resolution by downsampling the fMRI data to 0.25 Hz (4 seconds between timepoints).
We then fine-tuned new models on these downsampled fMRI responses as before.

\begin{figure}[t]
    \centering
    \includegraphics[width=1\linewidth]{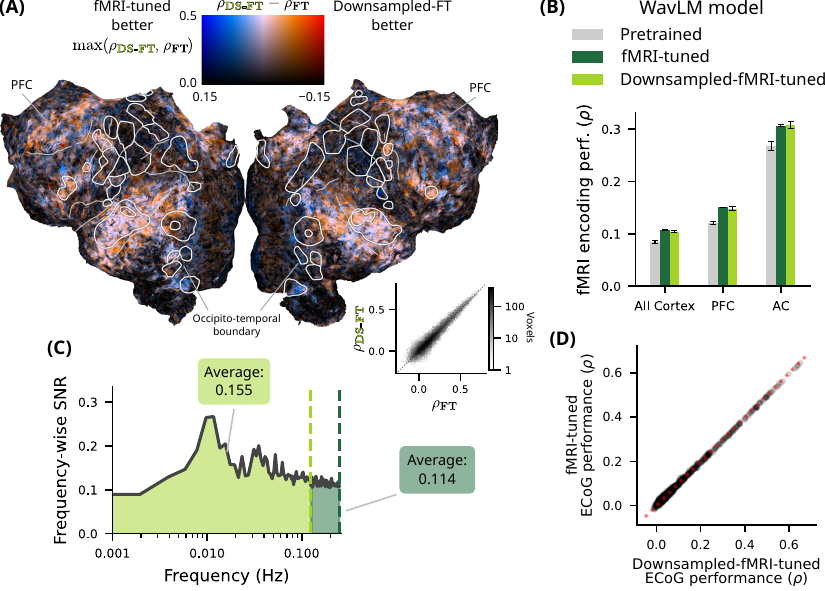}
    \caption{\textbf{Fine-tuning on downsampled fMRI responses.} (\textbf{a}) Flattened cortical surface of fMRI subject S3 that compares the encoding performance of models fine-tuned on original fMRI responses (blue) and downsampled fMRI responses (red). Voxel brightness is proportional to overall model performance. The 2-dimensional histogram shows that voxels have similar performance across fine-tuning conditions. We show subjects S1 and S2 in Appendix~\ref{app:downsampling-fmri-subjects}. (\textbf{b}) fMRI encoding performance, averaged within ROIs, after fine-tuning on original or downsampled fMRI responses. Error bars show standard error across subjects. There was no significant effect between the two-finetuning conditions (paired $t$-test across subjects had $p$-values 0.424, 0.565, and 0.798 for each respective ROI). (\textbf{c}) Signal-to-noise ratio (SNR) of the fMRI responses, averaged across subjects. The dark and light green vertical lines are the Nyquist frequency of the original and downsampled fMRI responses at 0.25 Hz and 0.125 Hz, respectively. The average SNR is $0.155 \pm 0.0143$ below the 0.125 Hz threshold and $0.114 \pm 0.000534$ above it. The original signal had an overall SNR of $0.139 \pm 0.00895$. This shows why downsampling the fMRI data before fine-tuning had little effect on model performance: most of the responses between \qtyrange[range-units=single,range-phrase=~--~]{0.125}{0.25}{\Hz} are noise. (\textbf{d}) Effect on ECoG performance after fine-tuning on downsampled or original fMRI responses. We see no significant change in performance due to fine-tuning data ($p=0.514$, two-sided paired $t$-test across all electrodes).}
    \label{fig:downsampled_ft}
\end{figure}

First, we compared how well the original fMRI-tuned and downsampled models predict held-out fMRI data. 
Figure~\ref{fig:downsampled_ft}a shows the difference in model prediction performance in one fMRI subject (S3).
We found that models fine-tuned on downsampled responses perform similarly to the original models overall, with slight decreases in performance in some higher-level areas like prefrontal cortex (PFC) and the boundaries of the occipital and temporal lobes.
Overall prediction performance was comparable, and both models outperformed the pretrained model in all brain areas (Figure~\ref{fig:downsampled_ft}b).

While surprising, this result can be explained by the signal-to-noise ratio (SNR) properties of fMRI responses to natural stimuli.
To explore this idea, we measured the SNR of the fMRI data across different frequency bands using a spectral approach \cite{hsuQuantifyingVariabilityNeural2004a} (see Section~\ref{sec:methods-snr}).
This showed that all frequency bands of the fMRI response are not equally informative. Most of the usable signal lies in the band between 0.01 and 0.1 Hz, while very little falls in the highest frequencies (Figure~\ref{fig:downsampled_ft}c).
Thus, by downsampling and truncating the noisiest frequencies, we actually \emph{raise} the overall SNR of the fMRI responses.
Fine-tuning on these ``cleaner'' downsampled responses allows the representations to generalize to the original signal.

Finally, we tested whether the models fine-tuned on downsampled fMRI data could still generalize to ECoG data. Figure~\ref{fig:downsampled_ft}d compares ECoG encoding model performance using fMRI-tuned and downsampled-fMRI-tuned models.
As was the case predicting fMRI data, we found virtually no difference between the two models at predicting ECoG data.
This demonstrates even more extreme temporal generalization than before: models fine-tuned on fMRI data with a Nyquist rate of 0.125 Hz improve performance on ECoG data.

\begin{figure}[t]
\centering
\includegraphics[width=1\linewidth]{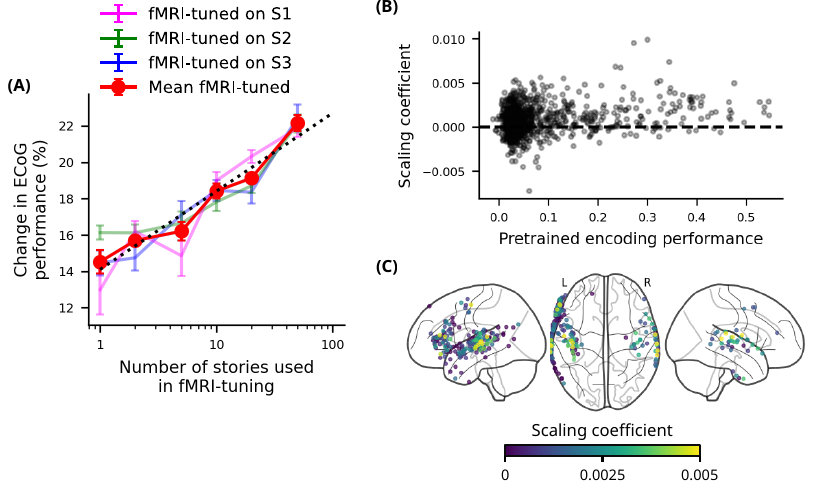}
\caption{\textbf{Scaling of fMRI-tuning for ECoG.} (\textbf{a}) ECoG encoding performance as a function of the number of fMRI fine-tuning stories. Error bars indicate the standard error over bootstraps of the fMRI stories. We show the scaling performance with the full fMRI-tuning dataset in Appendix~\ref{app:scaling-ecog-allstories}. (\textbf{b}) Pretrained encoding performance vs. scaling coefficient $m_e$ for all electrodes. The scaling law is measured per electrode as the slope of a linear regression between (1) its encoding performance and (2) log-fMRI story count. The electrodes that tend to improve have strong pretrained performance. (\textbf{c}) Electrodes were thresholded by pretrained encoding performance $\rho > 0.1$, yielding $n=219$ electrodes, and plotted on a brain atlas.}
\label{fig:scaling}
\end{figure}

\subsection{ECoG prediction scales with fMRI data}
\label{sec:results-scaling}

Several works have examined the scaling laws of encoding models in brain data~\cite{matsuyamaApplicabilityScalingLaws2023,gokceScalingLawsTaskOptimized2025}.
In particular, Antonello et al. (2023)~\cite{antonelloScalingLawsLanguage2023} showed that language encoding models in fMRI scale with logarithmically with the amount of training data.
Moussa et al. (2025)~\cite{moussaBraintuningImprovesGeneralizability2025} followed up this finding by fine-tuning the underlying models, and they found that fine-tuning reduces the amount of data needed to generalize to new subjects.
We next examined whether this effect holds across modalities as well.
To test scaling, we fine-tuned models using subsets of the LeBel et al. fMRI dataset, and then tested them on the ``Podcast'' ECoG dataset~\cite{antonelloScalingLawsLanguage2023}.

We found that improvements in overall ECoG prediction scaled logarithmically with the number of stories used for fMRI-tuning (Figure~\ref{fig:scaling}a).
This result mirrors the literature~\cite{antonelloScalingLawsLanguage2023,moussaBraintuningImprovesGeneralizability2025}, but with smaller improvement magnitudes due to transferring across modalities.
(In Appendix~\ref{app:scaling-fmri}, we additionally compare the scaling behavior of linear and fMRI-tuned models on fMRI encoding.)

We next quantified the scaling law of fMRI-tuning data on ECoG. Adapting the formulation of Antonello et al. (2023), for each electrode we fit a linear model $\Delta \rho_e \approx m_e \log_2 N$ to predict the change in encoding performance $\Delta \rho_e$ from the number of fMRI-tuning stories $N$. In Figure~\ref{fig:scaling}b, we show the scaling coefficient $m_e$ for all electrodes. Our trend was consistent across electrodes; after filtering for language-responsive electrodes, 176 of 219 electrodes across cortex had a positive relationship between encoding performance and the number of stories used for fMRI-tuning (Figure~\ref{fig:scaling}b).
Scaling effects were largest in areas with strong baseline encoding performance (Figure~\ref{fig:scaling}c).

Our results extend the literature by showing that fMRI-tuned models not only generalize across subjects, but also across recording methods. These results suggest that increasing amounts of fMRI data could be used to mitigate the limitations of intracranial data collection.

\section{Conclusion}
\label{sec:conclusion}

In this work, we showed that speech representations fine-tuned with fMRI data transfer to ECoG prediction, a recording modality with orders of magnitude higher temporal resolution. Transfer succeeds even when the fMRI data are further downsampled to 0.25 Hz, and performance scales logarithmically with the amount of fMRI fine-tuning data.

How is this possible? High-gamma power, the ECoG signal that we model, is thought to be the closest neural correlate to the BOLD signal measured in fMRI~\cite{logothetisNeurophysiologicalInvestigationBasis2001}. Thus it is plausible that fine-tuning on fMRI accentuates features that are relevant for high-gamma, even at response frequencies invisible to fMRI. This shared underpinning may be sufficient to enable slow-to-fast model transfer.

ECoG datasets are typically small: clinically necessary electrodes are implanted for a few days and then removed. fMRI datasets, in contrast, face no such constraints. Thus, as fMRI datasets continue to grow, our scaling results suggest that leveraging them will provide better and better ECoG encoding models. This is relevant for BCI applications where encoding model quality limits performance.

Still, several limitations apply to these results. We primarily focused on one model (WavLM Base+) and the single task domain of spoken language. Our approach is probably not fully data-optimal: there is also a rich landscape of training recipes involving nonlinear encoding adapters for ECoG data which we do not explore in this work, and there are other known means of fine-tuning acoustic representations to be more brain-like, such as fine-tuning on phonemes or semantic representations~\cite{vattikondaBrainWavLMFinetuningSpeech2025}. A wider search of this landscape would help clarify the optimal means by which fMRI data can be used to predict ECoG data. Yet it also seems likely that the optimal approach would combine all available data modes: fine-tune on fMRI \textit{and} ECoG \textit{and} other potential information sources in a joint training cocktail. Future work (and larger datasets) will be required to fully explore this space. Broadly, our work presents strong evidence that brain-like representations are highly modality invariant, and that out-of-modality training data can be a core component of the training mix for future brain encoding models.

\bibliographystyle{iclr2026_conference}
\bibliography{bib/fmri2ecog-neurips2026}

\appendix

\section{Per-subject change in ECoG performance after fMRI-tuning}
\label{app:ecog-per-subj}

\begin{figure}[H]
	\centering
	\includegraphics[width=1\linewidth]{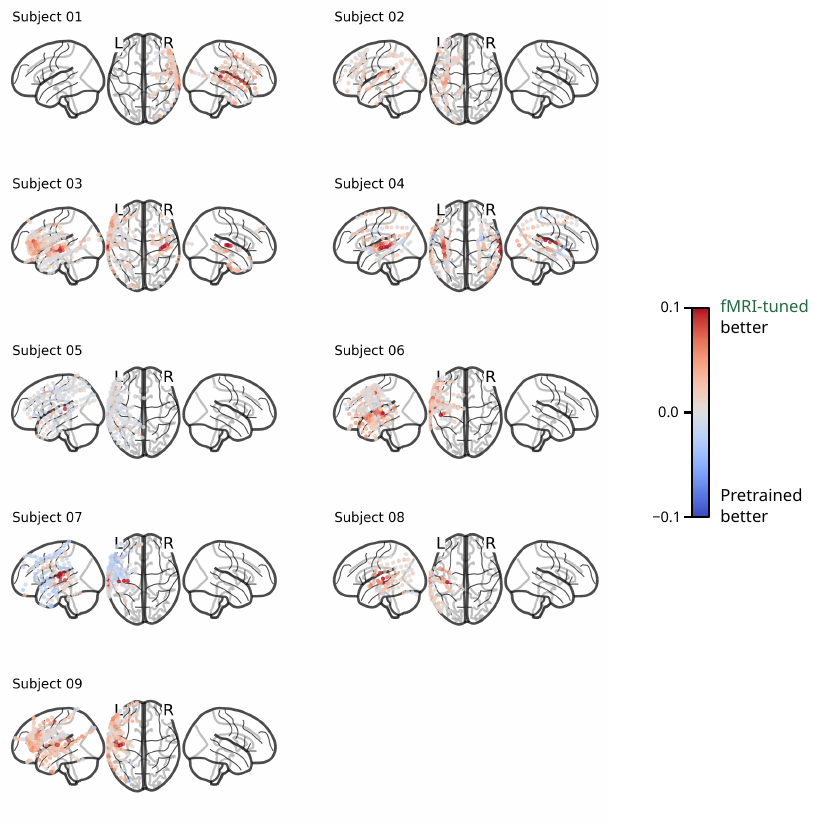}
	\caption{Improvement in encoding performance for all electrodes (Figure~\ref{fig:fmri2ecog_results}c), visualized separately for each subject.}
	\label{fig:app-ecog-per-subj}
\end{figure}

In Figure~\ref{fig:app-ecog-per-subj}, we visualize the per-electrode effect of fMRI-tuning separately for each subject in the ``Podcast'' dataset.

\newpage

\section{fMRI-tuning on downsampled fMRI responses}
\label{app:downsampling-fmri-subjects}

\begin{figure}[H]
	\centering
	\includegraphics[width=1\linewidth]{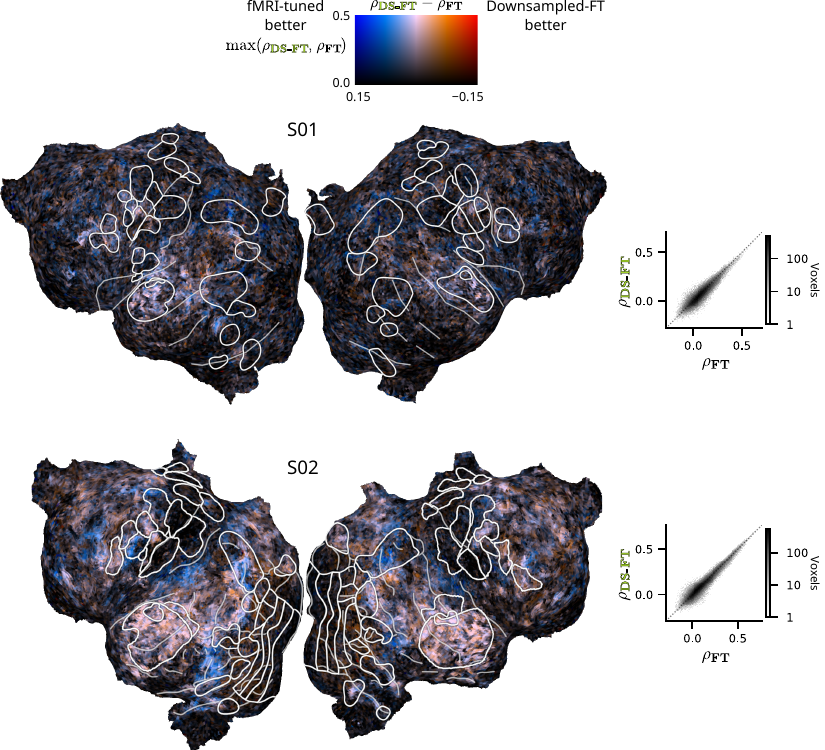}
	\caption{Flattened cortical surfaces of subjects S1 and S2 from LeBel et al. (2023)~\cite{lebelNaturalLanguageFMRI2023} that compare the encoding performance of the original fMRI-tuned models against the downsampled-fMRI-tuned models. On the right, 2-dimensional histograms show that individual voxel performance is stable across fine-tuning conditions. (See Figure~\ref{fig:downsampled_ft}a for S3.)}
	\label{fig:app-downsampling-fmri-subjects}
\end{figure}

Figure~\ref{fig:app-downsampling-fmri-subjects} compares the fMRI encoding performance of models fine-tuned on the original fMRI responses and the downsampled responses.

\newpage

\section{Scaling of fMRI-tuning for fMRI prediction}
\label{app:scaling-fmri}

\begin{figure}[H]
	\centering
	\includegraphics[width=1\linewidth]{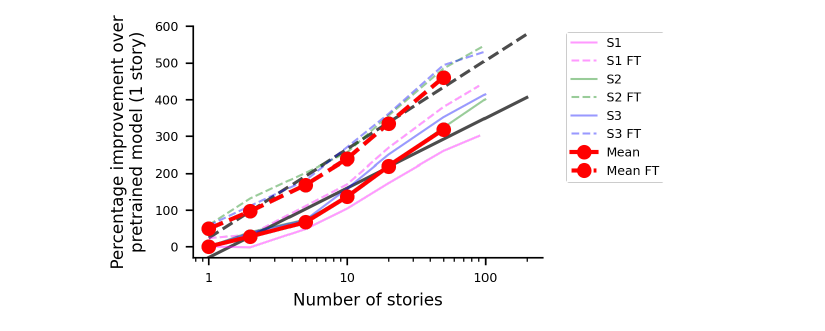}
	\caption{Within-subject fMRI encoding performance scales with the size of the fine-tuning dataset. Pink, green, and blue show scaling for each fMRI subject, and red shows the mean of the three. Solid lines show the scaling performance of the pre-trained model, while dashed lines show scaling of fine-tuned models. Black lines show a best linear fit to the mean across subjects. Fine-tuned models outperform the pretrained model at all dataset sizes and also improve faster.}
	\label{fig:app-scaling-fmri}
\end{figure}

Using the models fMRI-tuned in Section~\ref{sec:results-scaling}, we show how within-modality prediction scales with fine-tuning dataset size in Figure~\ref{fig:app-scaling-fmri}. Fine-tuned models improve at a faster rate than the pre-trained model, and they show gains even at small dataset sizes.

\newpage

\section{Scaling of fMRI-tuning for ECoG prediction} %
\label{app:scaling-ecog-allstories}

\begin{figure}[h!]
	\centering
	\includegraphics[width=1\linewidth]{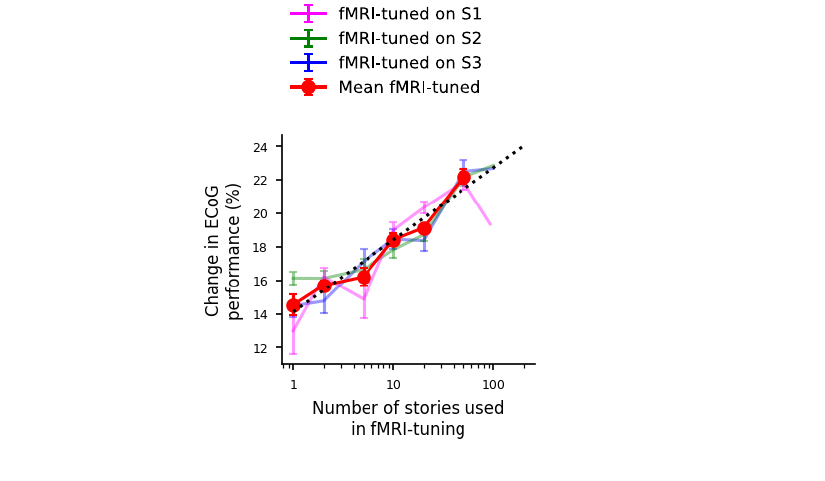}
	\caption{ECoG encoding performance as a function of the number of fMRI fine-tuning stories, including the full training set (93 stories for S1, and 100 stories for S2 and S3). Error bars indicate the standard error over bootstraps of the fMRI stories.}
	\label{fig:app-scaling-ecog-allstories}
\end{figure}

In Figure~\ref{fig:app-scaling-ecog-allstories}, we extend Figure~\ref{fig:scaling}a by showing fMRI-tuning performance using the full fMRI training set.

\end{document}